\pdfoutput=1
\documentclass[11pt]{article}

\usepackage{acl}

\usepackage{times}
\usepackage{latexsym}

\usepackage[T1]{fontenc}
\usepackage[utf8]{inputenc}

\usepackage{microtype}

\usepackage{inconsolata}

\usepackage{enumitem}
\usepackage{comment}
\usepackage{graphicx}
\usepackage{multirow}
\usepackage{bm}
\usepackage{xcolor}
\usepackage{amsmath}
\usepackage{amsfonts}
\usepackage{trimclip}
\usepackage{subcaption}
\usepackage{float}
\usepackage{listings}
\newcommand{\eg}{\textit{e.g.}}
\newcommand{\ie}{\textit{i.e.}}
\newcommand{\sft}{\textsc{sft}}
\newcommand{\icl}{\textsc{icl}}

\newcommand{\falcont}{\textsc{Falcon}-$7$\textsc{b}}
\newcommand{\falcons}{\textsc{Falcon}-$40$\textsc{b}}
\newcommand{\falconb}{\textsc{Falcon}-$180$\textsc{b}}
\newcommand{\llamatwochats}{\textsc{Llama-}$2$-$13$\textsc{b-Chat}}
\newcommand{\llamatwochatsshort}{\textsc{Llama}$2$-C}
\newcommand{\llamatwochatb}{\textsc{Llama-}$2$-$70$\textsc{b-Chat}}
\newcommand{\mixtral}{\textsc{Mixtral-8x7B-Instruct-v0.1}}
\newcommand{\mixtralshort}{\textsc{Mixtral}-I}
\newcommand{\lm}{\textsc{lm}}
\newcommand{\llm}{\textsc{llm}}
\newcommand{\flanultwo}{\textsc{Flan-ul}$2$-$20$\textsc{b}}
\newcommand{\boldflanultwo}{\textsc{\textbf{Flan-ul}}$\bm{2}$\textbf{-}$\bm{20}$\textbf{\textsc{b}}}
\newcommand{\qa}{\textsc{qa}}

\newcommand{\quac}{\textsc{q}u\textsc{ac}}
\newcommand{\doqa}{\textsc{d}o\textsc{qa}}
\newcommand{\chot}{\textsc{c}o\textsc{t}} % \cot didn't work
\newcommand{\schot}{\textsc{sc}o\textsc{t}}
\newcommand{\qlora}{\textsc{ql}o\textsc{ra}}

\newcommand{\uustate}{$uu$}
\newcommand{\austate}{$au$}
\newcommand{\acstate}{$ac$}
\newcommand{\ssstate}{$ss$}
\newcommand{\bolduustate}{$\bm{uu}$}
\newcommand{\boldaustate}{$\bm{au}$}
\newcommand{\boldacstate}{$\bm{ac}$}
\newcommand{\boldssstate}{$\bm{ss}$}
\newcommand{\shorttextrightarrow}{\clipbox*{{.5\width} 0pt {.95\width} {1.25\height}} \textrightarrow}
\newcommand{\twostatetrans}{\uustate{}\,\shorttextrightarrow{}\,\austate{}}
\newcommand{\boldtwostatetrans}{\bolduustate{}\,\shorttextrightarrow{}\,\boldaustate{}}
\newcommand{\threestatetranswithac}{\uustate{}\,\shorttextrightarrow{}\,\acstate{}\,\shorttextrightarrow{}\,\austate{}}

\newcommand{\fourstatetrans}{\uustate{}\,\shorttextrightarrow{}\,\acstate{}\,\shorttextrightarrow{}\,\ssstate{}\,\shorttextrightarrow{}\,\austate{}}
\newcommand{\boldthreestatetranswithac}{\bolduustate{}\,\shorttextrightarrow{}\,\boldacstate{}\,\shorttextrightarrow{}\,\boldaustate{}}
\newcommand{\boldfourstatetrans}{\bolduustate{}\,\shorttextrightarrow{}\,\boldacstate{}\,\shorttextrightarrow{}\,\boldssstate{}\,\shorttextrightarrow{}\,\boldaustate{}}

\definecolor{myGreen}{rgb}{0,.75,0}
\definecolor{myRed}{rgb}{.75,0,0}

\newlist{myitemize}{description}{10}
\setlist[myitemize]{labelindent=4pt, leftmargin=0pt, align=left, noitemsep, topsep=0pt}

\title{Structured Chain-of-Thought Prompting for\\Few-Shot Generation of Content-Grounded QA Conversations}

\author{Md Arafat Sultan\quad Jatin Ganhotra\quad Ramón Fernandez Astudillo\\ IBM Research AI\\ \texttt{\{arafat.sultan,ramon.astudillo\}@ibm.com}\\ \texttt{jatinganhotra@us.ibm.com}}

\begin{document}
\maketitle
\begin{abstract}
We introduce a structured chain-of-thought (\schot{}) prompting approach to generating content-grounded multi-turn question-answer conversations using a pre-trained large language model (\llm{}).
At the core of our proposal is a structured breakdown of the complex task into a number of states in a state machine, so that actions corresponding to various subtasks, \eg{}, content reading and utterance generation, can be executed in their own dedicated states.
Each state leverages a unique set of resources including prompts and (optionally) additional tools to augment the generation process.
Our experimental results show that \schot{} prompting with designated states for hallucination mitigation increases agent faithfulness to grounding documents by up to $16.8\%$.
When used as training data, our open-domain conversations synthesized from only $6$ Wikipedia-based seed demonstrations train strong conversational \qa{} agents; in out-of-domain evaluation, for example, we observe improvements of up to $13.9\%$ over target domain gold data when the latter is augmented with our generated examples.

\end{abstract}

\section{Introduction}
\label{section:introducion}

Despite enormous advances in large language models (\llm{}s) in recent years, their notorious propensity to hallucinate, \ie{}, generate text that are factually
% incorrect
inconsistent with existing knowledge, remains a critical issue \cite{xu2023understanding,bang2023multitask,huang2023survey}.
Of special interest to us in this paper are what \citet{openai2023gpt4} characterize as \textit{closed-domain} hallucinations, whereby models fail to generate text that can be supported by a given document even when they are explicitly instructed to do so \cite{xu2023understanding,qiu2023detecting,maynez2020faithfulness}.
Despite being powerful generators, \llm{}s can thus be unreliable readers, especially in the absence of extensive instruction tuning.

\begin{figure}[ht]
    \centering
    \includegraphics[scale=.48]{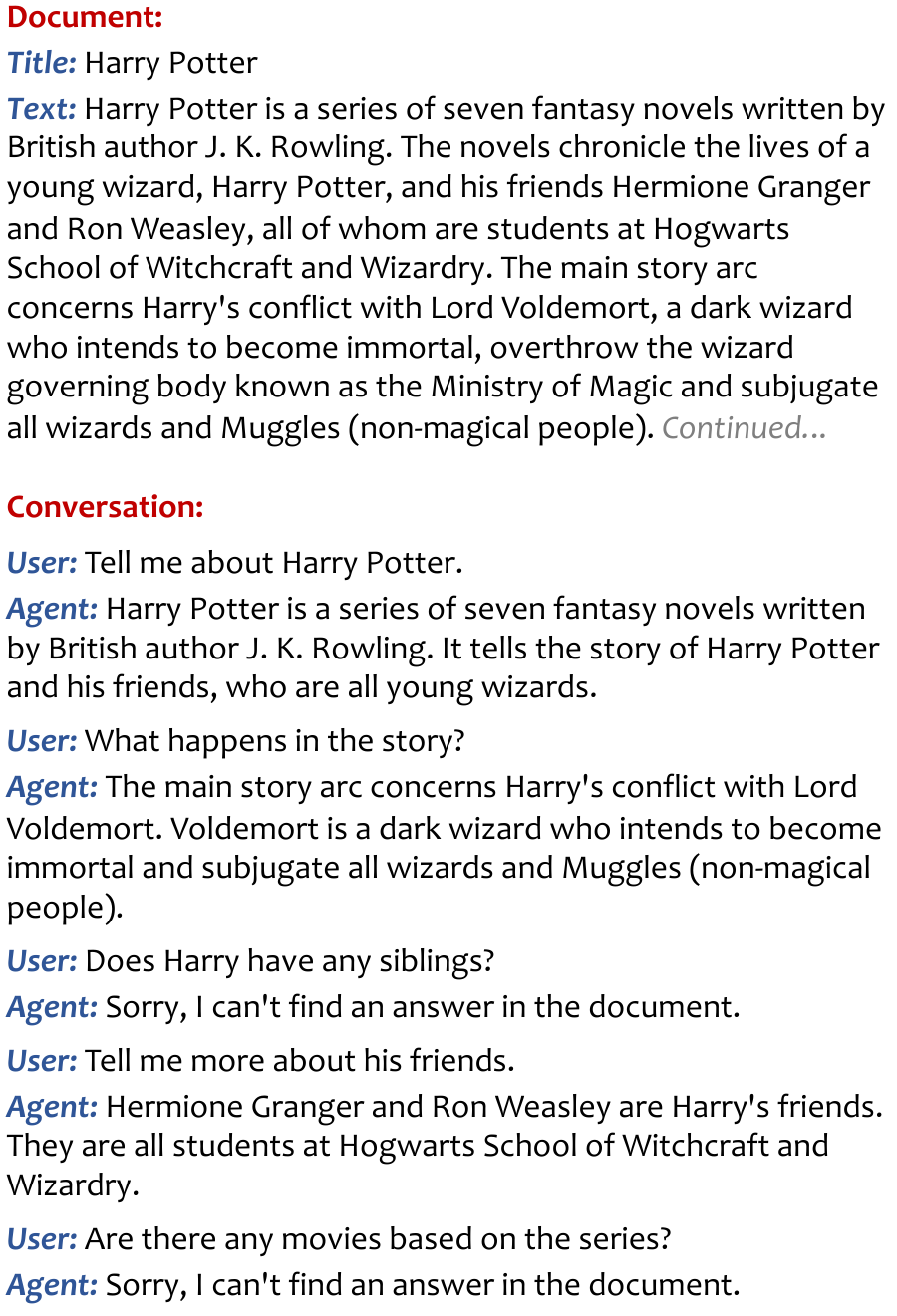}
    \caption{A multi-turn \qa{} conversation grounded in a document. If the document does not have an answer to a user query, the agent acknowledges so in its response.}
    \label{figure:harry-potter-demo}
    \vspace{-2mm}
\end{figure}

This paper studies closed-domain hallucination in pre-trained \llm{}s in the context of conversational question answering (\qa{}) \cite{choi2018quac,reddy2019coqa,adlakha2022topicocqa}, and its mitigation through inference-time augmentation.
Concretely, given a document, we want to generate a multi-turn \qa{} conversation between a \textit{user} and an \textit{agent}, in which the agent's responses to the user's queries must be grounded in the document.
%the user asks a series of related questions about the general topic of the document and the agent responds to those using information provided in the document.
%User queries have to make sense only in the context of the ongoing conversation and not on their own, \eg{}, due to pronominal references or as a follow-up to the previous query.
The requirements also crucially include for the agent to be able to determine if the given document has an answer to a query, and refrain from
%hallucinating one if not.
giving a memorized or made-up answer if not.
Figure~\ref{figure:harry-potter-demo} shows an example.
Compared to a single-turn setting \cite{mobashir2023delucionqa}, multi-turn \qa{} provides a more useful backdrop for studying \llm{} hallucinations, as an increased number of diverse user queries on a shared topic is likely to induce agent hallucinations more than a single \qa{} turn.

%We present novel algorithms for the task (\S{\ref{section:methodology}}) that operate by prompting a vanilla \llm{} using inexpensive few-shot in-context learning (\icl{}) \textcolor{red}{[REF]}.
We propose novel algorithms for this task (\S{\ref{section:methodology}}) that adhere to the general notion of language model (\lm{}) augmentation \cite{yao2023react,xu2023rewoo,schick2023toolformer}, whereby a set of actions -- each leveraging its own dedicated tools and resources -- are executed in an interleaved manner to solve a complex problem.
For example, besides generating utterances, we execute intermediate tasks that aim to answer questions such as: ``Is this user query answerable from the given document?'' or ``Where in the document is the answer?''
All subtasks are performed using few-shot in-context learning (\icl{}) \cite{brown2020language}, for which we utilize task-specific instructions, exemplars and (optionally) supporting models.

Our proposed actions and their execution sequences can be collectively represented as a state machine (\S{\ref{section:preliminaries}}), whose state transitions define our different algorithms (\S{\ref{section:methodology}}).
We refer to this approach as structured chain-of-thought (\schot{}) prompting, which, like ordinary chain-of-thought (\chot{}) prompting \cite{wei2022chain}, computes a final output through a set of careful reasoning steps, but unlike \chot{}, distributes those steps across designated states of a state machine.
A key aspect of our proposal is the simplicity of our individual actions, for which we are able to successfully prompt relatively small and open-source \llm{}s to generate high-quality conversations.

\begin{comment}
The algorithms can be collectively viewed as a state machine (Figure \ref{figure:state-machine}), whose states represent individual steps and sets of transitions through those states
%, along with the methods and resources they utilize, 
define entire algorithms (\S{\ref{section:preliminaries}}).
For example, our simplest algorithm executes only two steps: user and agent utterance generation, relying solely on instructions in its prompt to meet all task requirements.
A different, more advanced algorithm tries to mitigate agent hallucinations by explicitly assessing the answerability of a generated question from the given document in a third intermediate step, for which it can also (optionally) utilize a separate binary classifier.
State machines provide a natural framework for this complex task, offering finer-grained control over subtasks than generating the entire conversation in one go, and can be extended easily to incorporate new processes and resources through the addition of new states.
\end{comment}

Leveraging open-source models such as \falcons{} \cite{almazrouei2023falcon} and \flanultwo{} \cite{tay2023ul2} and a small set of Wikipedia-based exemplars that we create by hand (\S{\ref{section:methodology}}) -- the example of Figure~\ref{figure:harry-potter-demo} is one of them -- we generate open-domain \qa{} conversations from Wikipedia passages using our algorithms, and evaluate them both intrinsically and extrinsically (\S{\ref{section:experiments-and-results}}).
In intrinsic evaluation, we directly examine the quality of the generated conversations, including their faithfulness to the grounding document and their overall accuracy relative to pseudo-references provided by a high-performance instruction-tuned model \mixtral{} \cite{jiang2024mixtral}.
Results show that our proposed mechanisms for mitigating agent hallucinations do indeed reduce them by up to $16.8\%$, improving overall accuracy of agent utterances by as much as $7.7\%$.

In extrinsic evaluation, we train agents with our generated data to answer questions in multi-turn \qa{} conversations, and evaluate them against gold labels.
We report experiments with few-shot \icl{} and supervised fine-tuning (\sft{}) of agents on two conversational \qa{} datasets: \doqa{} \cite{campos2020doqa} and \quac{} \cite{choi2018quac}.
In \icl{} evaluation, our synthetic data -- generated from only $6$ seed demonstrations and with a relatively small \llm{} -- outperforms human-labeled data, including target domain gold data.
We also observe strong performance in \sft{} evaluation, which includes training conversational \qa{} agents only with synthetic data as well as augmenting existing target domain gold data with it.
For example, augmenting with our open-domain synthetic data improves agent performance over using only target domain gold data by an absolute $10$--$14\%$.

\begin{comment}
Our contributions can be summarized as follows:
\begin{itemize}[topsep=0pt,noitemsep,leftmargin=12pt]
    \item We present novel \textit{structured} chain-of-thought prompting methods with \lm{} augmentation for generating document-grounded \qa{} conversations with \llm{}s.
    \item In intrinsic evaluation, our proposed augmentations for mitigating hallucination help the \llm{} agent remain considerably more faithful to the given document.
    \item In extrinsic evaluation on grounded conversational \qa{} datasets, our generated conversations demonstrate strong standalone performance as well as the ability to effectively augment target domain gold data.
\end
{itemize}
\end{comment}

In summary, the following are our main contributions: (\textbf{\textit{i}})~We present novel \textit{structured} chain-of-thought prompting methods with \lm{} augmentation for generating document-grounded \qa{} conversations using \llm{}s; (\textbf{\textit{ii}}) In intrinsic evaluation, our proposed augmentations for hallucination mitigation help the \llm{} agent remain considerably more faithful to the given document; (\textbf{\textit{iii}})
In extrinsic evaluation on grounded conversational \qa{} datasets, our generated conversations demonstrate strong standalone performance as well as the ability to effectively augment target domain gold data.

\begin{figure*}[ht]
    \centering
    \includegraphics[scale=.45]{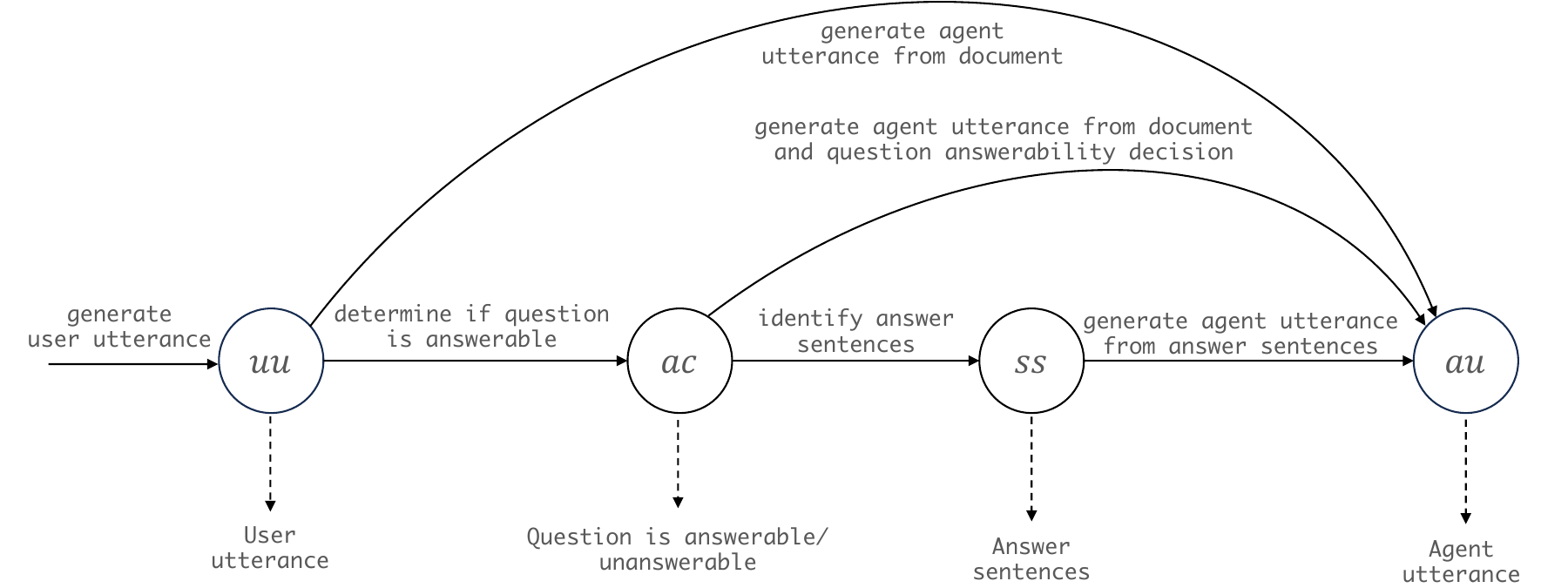}
    \caption{State machine for generating a single user-agent utterance pair within a multi-turn conversation (\S{\ref{section:preliminaries}}). An action (incoming arrow label) is executed in every state by few-shot prompting an \llm{} (\S{\ref{section:methodology}}), and an output is generated (dotted arrows). 
    One of multiple possible transitions then takes place (solid arrows), depending on the algorithm being run. 
    A grounding document and a conversation history (not in the diagram) are present in all steps.}
    \label{figure:state-machine}
\end{figure*}

\section{Preliminaries}
\label{section:preliminaries}
We introduce our state machine for \schot{} prompting in this section, and discuss the alignment of its states to different stages of generating a document-grounded multi-turn \qa{} conversation.
Further implementation details are provided in \S{\ref{section:methodology}}.

%Let $D$ be a random variable over the sample space $\mathbb{D}$ of all documents $d$, and $C_N$ the same over the set $\mathbb{C}_N$ of \qa{} conversations $\langle q_i,r_i \rangle_{i=1}^{N}$ of length $N$, where each $q_i$ is a user utterance (a query) and $r_i$ is a corresponding agent utterance (a response).
Let $\mathbb{D}$ be the set of all documents and $\mathbb{C}_N$ the set of \qa{} conversations $\langle q_i,r_i \rangle_{i=1}^{N}$ of length $N$, where each $q_i$ is a user utterance (a query) and $r_i$ is a corresponding agent utterance (a response).
%Let $D$ and $C_N$ be random variables corresponding to the sample spaces of $\mathbb{D}$ and $\mathbb{C}_N$, respectively.
Our task is to map an input document $d \in \mathbb{D}$ to a conversation $c_N=\langle q_1, r_1, ..., q_N, r_N \rangle \in \mathbb{C}_N$ such that $c_N$ is grounded in $d$: $c_N \sim p_{C_N \mid D}(\cdot \mid d)$.

\begin{comment}
\begin{figure*}[ht]
    \centering
    \begin{subfigure}{.5\textwidth}
    \centering
        \includegraphics[width=.9\linewidth]{au demo.pdf}
        \caption{Agent utterance generation (\boldaustate{}) with a pre-trained \llm{}}
        \label{subfigure:au-demo}
    \end{subfigure}%    
    \begin{subfigure}{.5\textwidth}
        \centering    \includegraphics[width=.9\linewidth]{ss demo.pdf}
        \caption{Answer sentence selection (\boldssstate{}) with an instruction-following model such as \flanultwo{}.}
        \label{subfigure:ss-demo}
    \end{subfigure}
    \caption{Example prompts for the states \boldaustate{} and \boldssstate{}. We only show 1-shot prompts here for simplicity; more demonstrations are used in practice.}
    \label{figure:icl-demos}    
\end{figure*}
\end{comment}

\begin{figure*}[ht]
    \centering
    \includegraphics[scale=.65]{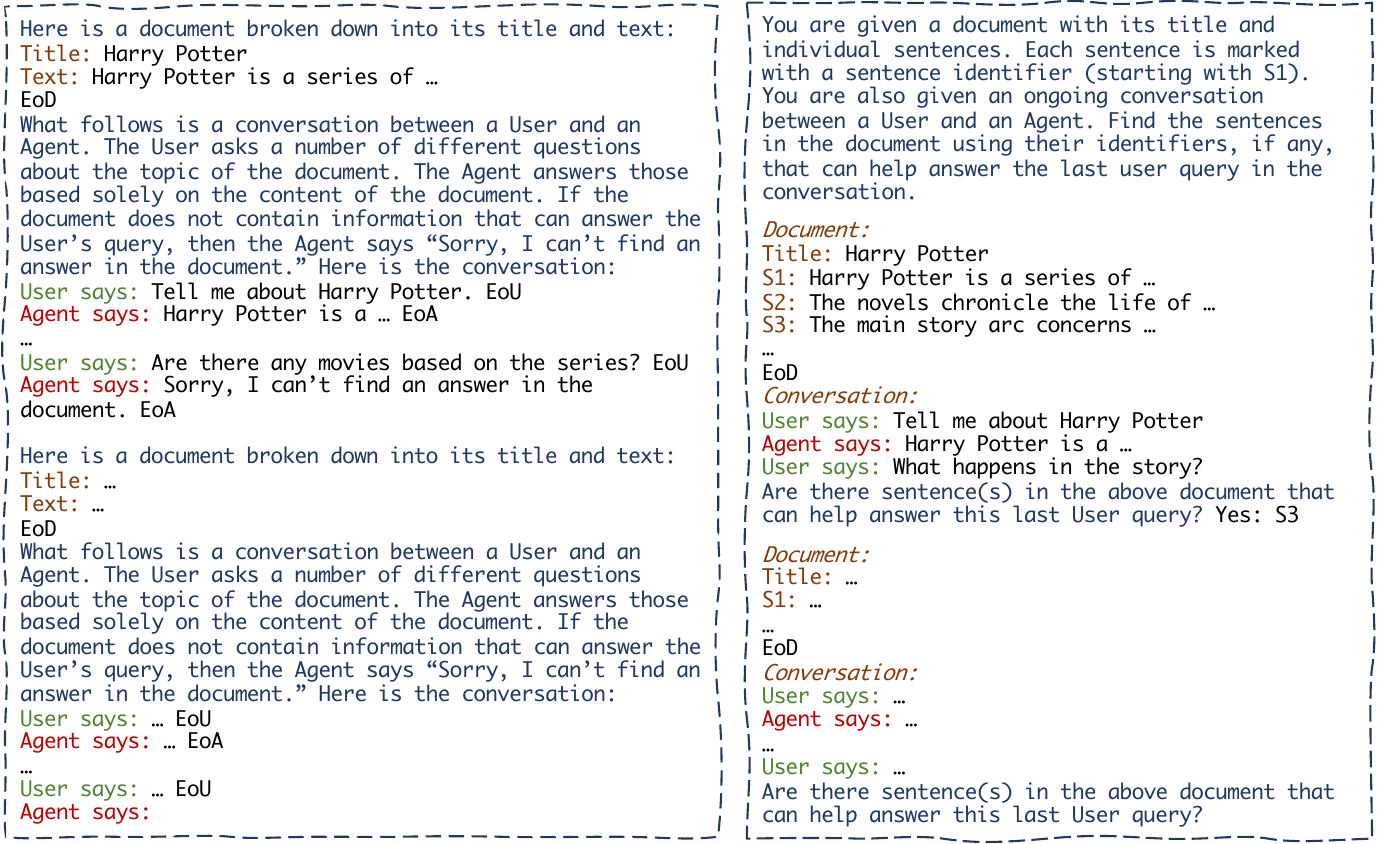}
    \caption{Prompts for states \boldaustate{} and \boldssstate{}. \textbf{Left:} Agent utterance generation (\boldaustate{}) with a pre-trained \llm{}. \textbf{Right:} Answer sentence selection (\boldssstate{}) with an instruction-following \llm{}.
    This diagram only shows $1$-shot prompts for brevity; we use more demonstrations in practice (see Appendix~\ref{appendix-section:prompts}).}
    \label{figure:au-ss-demos}
\end{figure*}

Our proposed algorithms prompt specific sets of state transitions in the state machine of Figure~\ref{figure:state-machine} to generate a single utterance pair $(q_i, r_i)$; a full conversation $c_N$ is generated by repeating the process $N$ times.
There are four states in the state machine, each corresponding to an individual action:
\begin{itemize}[topsep=1pt,itemsep=-.4ex,leftmargin=12pt]
    \item \textbf{User utterance generation (\bolduustate{})}: The next user utterance $q_i$ in an ongoing conversation $c_N$ is generated: $q_i \sim p_{uu}(\cdot \mid \langle q_j,r_j \rangle_{j=1}^{i-1}, d)$.
    
    \item \textbf{Question answerability classification (\boldacstate{})}: The current user query $q_i$ is classified as \textit{answerable} or \textit{unanswerable} from $d$: $a_i \sim P_{ac}(\cdot \mid q_i,\langle q_j,r_j \rangle_{j=1}^{i-1}, d)$.
    This state captures the notion that the assessment of the answerability of a user query can be kept separate from generating a response for it, which provides modularization and added flexibility. 
    For example, a different set of resources can now be leveraged for question answerability classification, such as a classifier trained on existing \qa{} data.
    
    \item \textbf{Answer sentence selection (\boldssstate{})}:
    Information pertaining to a single user query $q_i$ is often contained within a subset of all sentences in $d$.
    Before generating an agent response, it may be advantageous to identify those sentences so that response generation can focus more or solely on them \cite{sun2023generative,adolphs2022reason}.
    In state \boldssstate{}, the current query $q_i$ is mapped to relevant sentences in $d$: $s_1, ..., s_M \sim P_{ss}(\cdot \mid q_i,\langle q_j,r_j \rangle_{j=1}^{i-1}, d)$.
    %It is worth pointing out here that \boldssstate{} and \boldacstate{} represent steps that are closely related to chain-of-thought (\chot{}) prompting \textcolor{red}{[REFs]}, but are performed in a structured way in our proposed methods, \ie{}, inside designated states of a state machine.

    \item \textbf{Agent utterance generation (\boldaustate{})}: A response to the current query $q_i$ is generated:
    $r_i \sim p_{au}(\cdot \mid q_i,\langle q_j,r_j \rangle_{j=1}^{i-1}, d^*)$.
    State \boldaustate{} can be reached from any of the other three states; the following are to be noted: (\textit{i})~If \boldaustate{} is reached via a \boldtwostatetrans{} transition, then the input document $d^*=d$; (\textit{ii})~If \boldaustate{} is reached via an \boldssstate{}\,\shorttextrightarrow{}\,\boldaustate{} transition, then $d^*$ may contain only the sentences of $d$ deemed relevant to $q_i$ in \boldssstate{}, or alternatively special symbols marking those sentences; and finally, (\textit{iii}) If \boldaustate{} is reached via an \boldacstate{}\,\shorttextrightarrow{}\,\boldaustate{} transition, then this step is deterministic whenever $q_i$ is deemed unanswerable in \boldacstate{}, in which case $r_i$ takes the form of a pre-defined \textit{no answer} text.
\end{itemize}

Our methods can each be completely specified using (\textit{i}) the state transitions it executes in the state machine, and (\textit{ii}) the resources, \eg{}, models and associated prompts, that it utilizes in its different states, as we detail in \S{\ref{section:methodology}}.

\begin{comment}
\begin{enumerate}
    \item An algorithm consists of (\textit{a}) a generator, (\textit{b}) a set of states along with their inputs, transitions and outputs, and (\textit{c}) (optionally) a separate instruction-tuned assistant \llm{} for intermediate states.
    %\item Talk about roundtrip consistency test. 
    %\item Introduce the `Faithful Question Answering Agent' (or `\fqa{} Agent') role for \icl{}-based evaluation.    
\end{enumerate}
\end{comment}

\section{Methods}
\label{section:methodology}
We implement five algorithms that execute three unique sequences of state transitions.
What follows is a description of the sequences along with how the different algorithms implement them.
\begin{itemize}[topsep=1pt,itemsep=-.4ex,leftmargin=12pt]
    \item \boldtwostatetrans{}: One of our algorithms executes this transition to simply generate a user utterance first and then a corresponding agent utterance, both using a pre-trained \llm{}.
    The \llm{} is prompted with state-specific prefixes and manually created exemplars.
    Figure~\ref{figure:au-ss-demos} shows an example 1-shot prompt for state \boldaustate{} on the left; we use 2 exemplars in both states in practice.
    This prompt poses the task simply as one of text completion, which is what a pre-trained \llm{} is trained for.
    The prompt for \bolduustate{} works similarly, as shown in Figure~\ref{figure:uu-demo} of Appendix~\ref{appendix-section:prompts}.

    \item \boldthreestatetranswithac{}: We implement two algorithms that execute this sequence.
    Both generate an actual agent response in state \boldaustate{} only if the user's question is deemed answerable in state \boldacstate{}, otherwise the response is a fixed pre-defined string that indicates no answer.
    Both algorithms use a pre-trained \llm{} in states \bolduustate{} and \boldaustate{}.
    However, the two differ in their implementation of \boldacstate{}, for which one utilizes the same \llm{} and the other leverages a separate classifier fine-tuned on pre-existing \qa{} data.
    Separate prompts containing both answerable and unanswerable exemplars are used for the two types of models, which we show in Figures~\ref{figure:ac-demo} and \ref{figure:ac-plm-demo} of Appendix~\ref{appendix-section:prompts}.

    \item \boldfourstatetrans{}: Finally, two of our algorithms execute all four steps of the state machine.
    Upon classifying a question as answerable in state \boldacstate{}, relevant sentences are selected from the grounding document in state \boldssstate{}, which is provided as input to the step of \boldaustate{}.
    One of the algorithms utilizes the same pre-trained \llm{} in all four states; the other uses an instruction-tuned model in states \boldacstate{} and \boldssstate{} as these two states correspond to actions that are more akin to classification tasks.
    Figure~\ref{figure:au-ss-demos} shows a prompt for \boldssstate{} that instructs a \flanultwo{} model to select sentences using their identifiers.
\end{itemize}

The above methodology enables us to study the reading and generation aspects of our task separately and in a controlled manner, and better understand where pre-trained \llm{}s need the most assistance.
For example, we can explore the following questions directly: (\textit{a}) Is it useful to break down the task into reading (\boldacstate{}, \boldssstate{}) and generation (\bolduustate{}, \boldaustate{}) stages where \schot{} prompting can be utilized? and (\textit{b}) In the reading and reasoning states of \boldacstate{} and \boldssstate{}, can a pre-trained \llm{} perform the corresponding tasks by itself, or is a different set of tools needed?

\section{Experiments}
\label{section:experiments-and-results}
We write $6$ simple Wikipedia-based \qa{} conversations with a total of $32$ user turns and an equal number of agent turns; $20$ of the user queries are answerable and $12$ are unanswerable from their respective passages.
The example of Figure~\ref{figure:harry-potter-demo} is representative of the distribution; other titles include ``Table Tennis'' and ``Evolution''.
We then prompt \llm{}s with these as exemplars to synthesize new data from additional Wikipedia passages.

In this section, we first analyze and evaluate our generated datasets intrinsically.
To assess their practical utility, we then train conversational \qa{} agents on each and evaluate on different test sets.
As stated before, each generation algorithm is uniquely specified by its three components: (\textit{a})~the \textbf{\textit{generator:}} we use \falcons{} to generate all user and agent utterances; (\textit{b})~the \textbf{\textit{state transition sequence:}} the sequence of state transitions taken by the algorithm in the state machine; and (\textit{c})~the \textbf{\textit{assistant:}} some algorithms additionally use a \flanultwo{} model in states \boldacstate{} and \boldssstate{} (if applicable) to augment generation. 
We use nucleus sampling ($p$$=$$.9$) \cite{holtzman2023curious} for user utterance generation and greedy decoding in all other steps.
Details of model prompting at different stages of generation are provided in Appendix~\ref{appendix-section:prompts}.

\begin{table*}[h]
    \scriptsize
    \centering
    \begin{tabular}{c|c|c|c|c|c|c|c|c|c|c|c}
        \multirow{2}{*}{\textbf{\shortstack[c]{State\\Transitions}}} & \multirow{2}{*}{\textbf{Assistant}} & \multirow{2}{*}{\shortstack[c]{$\bm{\%}$\\\textbf{Has Answer}}} & \multirow{2}{*}{\shortstack[c]{$\bm{\%}$\\\textbf{Extracted}}} & \multicolumn{2}{c|}{\textbf{Faithfulness}} & \multicolumn{3}{c|}{\textbf{Cls-Acc-\mixtralshort{}}} & \multicolumn{3}{c}{\textbf{F1-\mixtralshort{}}} \\
        %\cline{5-12}
        & & & & \textbf{Lexical} & \textbf{WeCheck} & \textsc{a} & \textsc{ua} & \textsc{hm} & \textsc{a} & \textsc{ua} & \textsc{hm} \\
        \hline
        \twostatetrans{} & \textsc{n/a} & $74.9$ & $19.0$ & $83.3$ & $71.8$ & $87.8$ & $44.0$ & $58.6$ & $46.3$ & $44.0$ & $45.1$ \\
        %\hline
        \cline{1-2}
        \multirow{2}{*}{\threestatetranswithac{}} & None & $69.6$ & $17.3$ & $83.8$ & $72.8$ & $82.4$ & $49.4$ & $61.8$ & $43.0$ & $49.4$ & $46.0$ \\
        %\cline{2-12}
        & \flanultwo{} & $47.6$ & $20.9$ & $93.6$ & $86.5$ & $68.7$ & $85.9$ & $76.3$ & $37.8$ & $85.9$ & $52.7$ \\
        %\hline
        \cline{1-2}
        \multirow{2}{*}{\fourstatetrans{}} & None & $58.4$ & $58.5$ & $90.5$ & $80.6$ & $67.1$ & $55.5$ & $60.7$ & $30.7$ & $55.5$ & $39.5$ \\
        %\cline{2-12}
        & \flanultwo{} & $51.9$ & $50.8$ & $96.5$ & $88.6$ & $74.2$ & $85.6$ & $79.5$ & $38.2$ & $85.6$ & $52.8$\\
    \end{tabular}
    \caption{Key statistics and \llm{}-as-reference evaluation results for data synthesized by a few-shot prompted \falcons{} model, optionally assisted by a few-shot prompted \flanultwo{} model in states \boldacstate{} and \boldssstate{} (if applicable). Accuracy of classification into answerable/unanswerable classes (Cls-Acc) and F1-scores are computed against pseudo-references provided by a prompted \mixtral{} model. \textsc{a/ua}:~the class of answerable/unanswerable questions; \textsc{hm}:~the harmonic mean of the two previous columns.}
    \label{table:intrinsic-analysis}
\end{table*}

\subsection{Intrinsic Evaluation and Analysis}
\label{subsection:intrinsic-eval}
We sample $1,000$ Wikipedia passages and generate a conversation from every passage using each of our five algorithms; every conversation consists of $5$ user and $5$ agent utterances.
Table~\ref{table:intrinsic-analysis} shows the results of our intrinsic analysis of this data.
The first two columns provide two basic statistics: the $\%$ of user queries that are responded with an answer (as opposed to \textit{no answer}) and the $\%$ of answers that are completely extracted from the given passage, \ie{}, without any abstraction.
We observe that: (\textit{a})~algorithms that transit through states \boldacstate{} and \boldssstate{}, \ie{}, those that explicitly reason about query answerability and answer sentences, deem more questions as unanswerable, with the \flanultwo{} assistant predicting thus more often than \falcons{}, and (\textit{b})~algorithms that search for relevant sentences (\boldssstate{}) often copy document sentences verbatim in the answer, exhibiting the least abstraction.
%We will see later how these properties relate to performance.

Next we evaluate the faithfulness of the generated agent answers to the grounding document using two metrics: (\textit{a})~the lexical precision of the answer with respect to the document after stopword removal and stemming, and (\textit{b})~the WeCheck \cite{wu2023wecheck} factual consistency score of the same.
Note that this evaluation only concerns actual answers and not the \textit{no answer} responses, as we aim to exclusively measure hallucination with it.
The results of Table~\ref{table:intrinsic-analysis} clearly indicate that both answerability classification (\boldacstate{}) and answer sentence selection (\boldssstate{}) improve agent faithfulness through \schot{} prompting, and use of the \flanultwo{} assistant leads to more faithful generation than using \falcons{} in these steps.
Overall, we observe an improvement of up to $16.8\%$ over the simplest algorithm defined by the \boldtwostatetrans{} transition.

\begin{table*}[h]
    \centering
    \footnotesize
    \begin{tabular}{ccc|ccc}
        \multirow{2}{*}{\textbf{Generator}} & \multirow{2}{*}{\shortstack[c]{\textbf{State}\\\textbf{Transitions}}} & \multicolumn{1}{c|}{\multirow{2}{*}{\textbf{Assistant}}} & \multirow{2}{*}{$\bm{F_1}$-\textbf{HM}}  & \multirow{2}{*}{\shortstack[c]{$\bm{F_1}$\\(A)}} & \multirow{2}{*}{\shortstack[c]{$\bm{F_1}$\\(UA)}} \\
        & & \multicolumn{1}{c|}{} & & & \\
        \hline
        %\cline{1-7}
        None (0-shot) & \textsc{n/a} & \textsc{n/a} & $26.1{\scriptstyle \pm4.5\%}$ & $54.0$ & $17.2$ \\
        %\hline
        \textsc{Human} & \textsc{n/a} & \textsc{n/a} & $46.6{\scriptstyle \pm5.5\%}$ & $67.3$ & $35.7$ \\
        \hline
        \multirow{5}{*}{\falcons{}} & \twostatetrans{} & \textsc{n/a} & $38.4{\scriptstyle \pm4.6\%}$ & $67.9$ & $26.8$ \\
        \cline{2-3}
        & \multirow{2}{*}{\boldthreestatetranswithac{}} & None & $41.0{\scriptstyle \pm2.9\%}$ & $67.5$ & $29.5$ \\
        %\cline{3-3}
        & & \boldflanultwo{} & $\bm{54.6^*}{\scriptstyle \pm2.5\%}$ & $64.5$ & $47.4$ \\
        \cline{2-3}
        & \multirow{2}{*}{\fourstatetrans{}} & None & $45.1{\scriptstyle \pm2.0\%}$ & $64.9$ & $34.6$ \\
        %\cline{3-3}
        & & \flanultwo{} & $52.4^*{\scriptstyle \pm3.6\%}$ & $63.7$ & $44.6$ \\
    \end{tabular}
    \caption{In-domain performance of \llamatwochats{} as the \qa{} Agent on our seed demonstrations when prompted with various datasets. Asterisks (*) indicate improvement over human-annotated data.}
    \label{table:icl-results-on-seed-demonstrations}    
\end{table*}

Finally, we take an \llm{}-as-a-reference approach to assessing the overall quality of our generated data as follows: We first prompt a high-performance \mixtral{} model (\mixtralshort{} from here on) to generate agent utterances for all user queries in our synthetic data.
\mixtralshort{} is given the queries along with their conversation histories, as generated by the original algorithm.
We then evaluate the original agent utterances against the ones generated by \mixtralshort{}.
We observed this evaluation strategy to be generally more reliable than asking \mixtralshort{} to judge generated utterances.
In essence, we measure the extent to which our different processes behave like a high-performance instruction-tuned \llm{}.
%the proximity of the output of our different algorithms 
%-- which use only a pre-trained \llm{} for generation --
%to those of a high-performance instruction-tuned \llm{}.

Importantly, the presence of both answerable and unanswerable questions in our task calls for an evaluation protocol that incorporates the two individual classes and also assesses holistically on both.
We measure lexical unigram precision and recall (after stopword removal and stemming) in the answerable class (as deemed by \mixtralshort{}) and compute a final F1-score; for the unanswerable class, a second F1-score is computed where precision and recall are both $1$ if the agent response indicates that the question is unanswerable, otherwise both are $0$.
We use the harmonic mean of the two F1-scores as our final evaluation metric to reward class-balanced performance.

The last three columns of Table~\ref{table:intrinsic-analysis} show the performances of all five algorithms.
The three and four-step \schot{}-prompted algorithms augmented with a \flanultwo{} assistant have the best combined scores.
On a closer look, the improvements from these approaches can be attributed to a big jump in performance in the unanswerable class, trading off much less accuracy in the answerable class.
We also look at the related metric of answerability classification accuracy (Cls-Acc in Table~\ref{table:intrinsic-analysis}) and observe a clear correlation with F1-score: algorithms with better overall performances are those that are better and more balanced at question answerability classification.
These results crucially suggest that the primary source of hallucination in \falcons{} in our data is misclassification, as it often produces answers to unanswerable questions.

\subsection{Extrinsic Evaluation}
\label{subsection:extrinsic-eval}

\subsubsection{Setup}
\label{subsubsection:extrinsic-eval-setup}
Our extrinsic evaluation involves training \qa{} agents with our generated data to produce responses to user queries in an ongoing conversation, and evaluating the agents on unseen test sets.
We use two conversational \qa{} datasets in our experiments: \doqa{} \cite{campos2020doqa} and \quac{} \cite{choi2018quac}.
%\textcolor{red}{Jatin: I believe we can move the rest of this para to Appendix to save space. @Arafat: Thoughts? If yes, say Yes and I'll move them.} \textcolor{blue}{@Jatin: Not yet, I feel we should provide some bare minimum stats in the main body on datasets, but we'll still consider it later}
\doqa{} (v2.1) has a training set of $1,037$ conversations containing $4,612$ dialogue turns; the test set has $1,200$ dialogues, with more than $4$ turns on average per dialogue in three different domains: Cooking, Travel and Movies.
For Wikipedia-based \quac{}, we use the official dev set of $1,000$ conversations as our test set, and split the official train set of $11,567$ conversations into $10,567$ for training and $1,000$ for validation.
\quac{} has $7.2$ turns per dialogue on average.
We refer to the original papers for more detailed statistics.

We evaluate all generated data using both few-shot prompting and supervised fine-tuning of \qa{} agents, as described in the next two sections.
Our evaluation metric is the F1-score of \S{\ref{subsection:intrinsic-eval}}.

\subsubsection{Few-Shot Prompting}
\label{subsubsection:extrinsic-eval-icl-results}

In few-shot evaluation, a trainee \qa{} agent receives inference-time supervision in the form of in-context learning (\icl{}) demonstrations, which are sampled from our generated data.
The agent is then asked to produce a response to a user query in an ongoing gold conversation based on a grounding document, similarly to state \boldaustate{} of our state machine (\S{\ref{section:preliminaries}}).
Given the limited amount of supervision that can be provided through \icl{}, we use an existing open-domain chat model \llamatwochats{} \cite{touvron2023llama} (\llamatwochatsshort{} henceforth) in our few-shot evaluation experiments.

\begin{table*}[ht]
    \footnotesize
    \centering
    \begin{subtable}{1\linewidth}
    \centering
    \begin{tabular}{ccc|ccc}
        \multirow{2}{*}{\textbf{Generator}} & \multirow{2}{*}{\shortstack[c]{\textbf{State}\\\textbf{Transitions}}} & \multicolumn{1}{c|}{\multirow{2}{*}{\textbf{Assistant}}} & \multirow{2}{*}{$\bm{F_1}$-\textbf{HM}}  & \multirow{2}{*}{\shortstack[c]{$\bm{F_1}$\\(A)}} & \multirow{2}{*}{\shortstack[c]{$\bm{F_1}$\\(UA)}} \\
        & & \multicolumn{1}{c|}{} & & & \\
        \hline
        %\cline{1-7}
        None (0-shot) & \textsc{n/a} & \textsc{n/a} &  $30.5\scriptstyle\pm0.0\%$ & $28.7$ &  $32.6$ \\
        %\hline
        \textsc{Human} (Wikipedia) & \textsc{n/a} & \textsc{n/a} & $46.0\scriptstyle\pm0.6\%$ & $38.3$ & $57.7$ \\
        %\hline
        \textsc{Human} (\doqa{}) & \textsc{n/a} & \textsc{n/a} & $46.2\scriptstyle\pm0.9\%$ & $49.6$ & $43.2$ \\
        %\hline
        \falcons{} (Wikipedia) & \fourstatetrans{} & \flanultwo{} & $\mathbf{50.4^*}\scriptstyle\pm0.9\%$ & $40.4$ & $66.8$ \\
        %\hline
        %\multirow{2}{*}{\falconb{}} & \twostatetrans{} & \textsc{n/a} & \falcons{} & $\bm{49.1}\scriptstyle\pm0.9\%$ & $41.6$ & $59.7$ \\
        %\cline{2-7}
        %& \fourstatetrans{} & \flanultwo{} & \llamatwochats{} & $\bm{48.1}\scriptstyle\pm0.5\%$ & $38.8$ & $63.1$ \\
        %\hline\
    \end{tabular}
    \caption{\textbf{\doqa{} (Cooking) Results}}
    \end{subtable}

    \vspace{2mm}
    
    \begin{subtable}{1\linewidth}
    \centering
    \begin{tabular}{ccc|ccc}
        %\multirow{2}{*}{\textbf{Generator}} & \multirow{2}{*}{\shortstack[c]{\textbf{State}\\\textbf{Transitions}}} & \multicolumn{1}{c|}{\multirow{2}{*}{\textbf{Assistant}}} & \multirow{2}{*}{$\bm{F_1}$-\textbf{HM}}  & \multirow{2}{*}{\shortstack[c]{$\bm{F_1}$\\(A)}} & \multirow{2}{*}{\shortstack[c]{$\bm{F_1}$\\(UA)}} \\
        %& & \multicolumn{1}{c|}{} & & & \\
        \hline
        %\cline{1-7}
        None (0-shot) & \textsc{n/a} & \textsc{n/a} & $39.2\scriptstyle\pm0.0\%$ & $44.4$ & $35.1$ \\
        %\hline
        \textsc{Human} (Wikipedia) & \textsc{n/a} & \textsc{n/a} & $53.6\scriptstyle\pm2.2\%$ & $51.9$ &  $55.4$ \\
        %\hline
        \textsc{Human} (\quac{}) & \textsc{n/a} & \textsc{n/a} & $50.1\scriptstyle\pm0.7\%$ & $56.5$ & $45.1$ \\
        %\hline        
        \falcons{} (Wikipedia) & \fourstatetrans{} & \flanultwo{} & $\mathbf{56.5^*}\scriptstyle\pm2.7\%$ & $52.4$ & $61.3$ \\
        %\hline
        %\multirow{2}{*}{\falconb{}} & \threestatetranswithac{} & \flanultwo{} & \falcons{} & %$\bm{49.6}\scriptstyle\pm2.2\%$ & $41.6$ & $61.7$ \\
        %\cline{2-7}
        %& \threestatetranswithac{} & $\times$ & \llamatwochats{} & $\bm{55.8}\scriptstyle\pm0.3\%$ & $50.7$ & $62.0$ \\
        %\hline
    \end{tabular}
    \caption{\textbf{\quac{} Results}}
    \end{subtable}
    \caption{Performance of \llamatwochats{} as a \qa{} Agent on two external benchmarks when few-shot prompted with various datasets. Data generated by our $4$-step algorithm outperforms human-written demonstrations.}
    \label{table:ood-icl-results}
\end{table*}

Our first \icl{} evaluation re-uses the $6$ seed demonstrations for an in-domain \textit{roundtrip} assessment as follows: Given a $1000$-dialogue synthetic dataset from \S{\ref{subsection:intrinsic-eval}} as the population of demonstrations, we run $c$ evaluation cycles, each consisting of $r$ rounds of evaluation.
In each round, we uniformly sample (\textit{a}) $2$ document-conversation pairs from our synthetic dataset, and (\textit{b}) one of the $32$ user queries from the seed demonstrations along with its grounding document and conversation history.
We then prompt \llamatwochatsshort{} with the synthetic demonstrations to generate an agent response for the seed query, and evaluate against the gold response.
An evaluation cycle is complete when all $r$ rounds in it have ended and the results have been averaged.
Our final results consist of an average F1-score over the $c$ cycles.
Table~\ref{table:icl-results-on-seed-demonstrations} shows evaluation results for $c=3$ and $r=1000$.
Standard deviations are reported as a $\%$ of the mean for comparability across methods.
We first observe that all $2$-shot results are better than the $0$-shot results, indicating that \llamatwochatsshort{} can benefit from runtime demonstrations.
Second, our methods that execute \schot{} prompting with dedicated actions for question answerability classification and answer sentence selection, especially with a \flanultwo{} assistant, perform the best.
As in \S{\ref{subsection:intrinsic-eval}}, they achieve a better balance between answerable and unanswerable class performance.
The human demonstrations are sampled from our seed data so that the test query is from a different conversation.
Interestingly, two of our methods outperform human demonstrations, likely due to the (much larger) synthetic datasets containing more topically similar passages with test instances.

Next we evaluate on the \doqa{} (Cooking) and \quac{} test sets.
The setup is similar as before, with one difference: instead of randomly sampling an instance from the test set in each round, we evaluate on all test instances to complete a cycle.\footnote{For \quac{} training, we are able to use only one conversation -- often longer than the other datasets -- as demonstration before the input length reaches the limit for \falcons{}.}
We report results from only the best-performing synthetic datasets in Table~\ref{table:ood-icl-results}.
The four-step algorithm with a \flanultwo{} assistant produces the best synthetic data for \icl{} on both test sets, again outperforming our seed demonstrations.
Interestingly, the synthetic data improves performance on the unanswerable class even over target domain human-labeled data, resulting in less hallucination and better overall results.

In summary, we observe in both in-domain and out-of-domain evaluation that \schot{} prompting methods that leverage a \flanultwo{} assistant to reduce hallucination perform the best for \icl{}.

\begin{table*}[h]
    \centering
    \footnotesize
    \begin{tabular}{l|ccc|ccc|ccc|ccc}
         \multicolumn{1}{c|}{\multirow{3}{*}{\textbf{Training Set}}} & \multicolumn{12}{c}{\textbf{Evaluation Sets}} \\
         \cline{2-13}
         & \multicolumn{3}{c|}{\textbf{\doqa{} Cooking}}
         & \multicolumn{3}{c|}{\textbf{\doqa{} Movies}} & \multicolumn{3}{c|}{\textbf{\doqa{} Travel}} & \multicolumn{3}{c}{\textbf{\quac{}}}\\
         & \textsc{a} & \textsc{ua} & \textsc{hm} & \textsc{a} & \textsc{ua} & \textsc{hm} & \textsc{a} & \textsc{ua} & \textsc{hm} & \textsc{a} & \textsc{ua} & \textsc{hm} \\
         \hline
         $D$: \doqa{} \textsc{Gold} & $26.4$ & $60.1$ & $36.6$ & $20.8$ & $71.1$ & $32.2$ & $25.1$ & $63.3$ & $35.9$ & $22.1$ & $73.1$ & $34.0$ \\
         $Q$: \quac{} \textsc{Gold} & $10.2$ & $90.6$ & $18.3$ & $7.1$ & $95.5$ & $13.2$ & $8.1$ & $97.2$ & $15.0$ & $35.0$ & $85.6$ & $49.7$ \\
         \hline
         $S_1$: \twostatetrans{} & $22.8$ & $65.2$ & $33.8$ & $19.0$ & $59.4$ & $28.8$ & $19.0$ & $67.0$ & $29.6$ & $38.6$ & $43.1$ & $40.7$\\
         $S_2$: \fourstatetrans{} & $19.2$ & $85.8$ & $31.3$ & $14.2$ & $90.4$ & $24.5$ & $18.1$ & $87.5$ & $29.9$ & $28.4$ & $82.8$ & $42.3$ \\
         $S_3$: $50\%$ $S_1$ $\cup$ $50\%$ $S_2$ & $22.5$ & $71.9$ & $34.3$ & $18.4$ & $77.1$ & $29.7$ & $18.9$ & $78.0$ & $30.5$ & $37.2$ & $49.3$ & $42.4$  \\
         \hline
         $D$ augmented w/ $S_3$ & $40.7$ & $66.3$ & $\mathbf{50.4}$ & $34.2$ & $66.4$ & $\mathbf{45.1}$ & $39.9$ & $66.2$ & $\mathbf{49.8}$ & \multicolumn{3}{c}{--} \\
         $Q$ augmented w/ $S_3$ & \multicolumn{3}{c|}{--} & \multicolumn{3}{c|}{--} & \multicolumn{3}{c|}{--} & $49.8$ & $75.0$ & $\mathbf{59.8}$ \\
    \end{tabular}
    \caption{Performances of \falcont{} models fine-tuned (with \qlora{}) on various training datasets (\S{\ref{subsubsection:extrinsic-eval-sft-results}}).}
    \label{table:sft-results}
\end{table*}

\subsubsection{Supervised Fine-Tuning (\sft{})}
\label{subsubsection:extrinsic-eval-sft-results}

To further examine the utility of our generated data, next we fine-tune a pre-trained \llm{} (\falcont{}) on each dataset with \qlora{} \cite{dettmers2023qlora}, and evaluate on the test sets of \doqa{} and \quac{}.
Given the relatively high cost of \sft{} experiments involving our suite of five algorithms, we adopt the following two-stage process for evaluation: 1.~Compare all five algorithms by (1\textit{a})~training a \falcont{} model on relatively small amounts of data from each and (1\textit{b})~evaluating on validation sets; and 2.~Assess the utility of the two best algorithms identified in step 1 more closely by (2\textit{a})~generating more data with each to fine-tune a new \falcont{} model and (2\textit{b})~evaluating on test sets.

In our implementation of step 1 (details in Appendix \ref{Appendix: SFT training details}), the five $1000$-conversation datasets of \S{\ref{subsection:intrinsic-eval}} are re-used as training data.
The evaluation identifies the simplest \boldtwostatetrans{} algorithm and the most advanced \boldfourstatetrans{} with a \flanultwo{} assistant as the two best algorithms on the two validation sets (see Table~\ref{table:sft-results-appendix-1000} of Appendix~\ref{Appendix: SFT training details}).
Interestingly, despite generating the least faithful conversations among our different algorithms (\S{\ref{subsection:intrinsic-eval}}), and unlike in few-shot prompting (\S{\ref{subsubsection:extrinsic-eval-icl-results}}), the two-step algorithm performs strongly in \sft{}, indicating that \sft{} can be more robust to noisy training data than \icl{} as long as there is useful signal in it.

For step 2, we generate $10,000$ conversations with each of the above two algorithms.
Table~\ref{table:sft-results} presents a detailed comparison of performances on all four \doqa{} and \quac{} test sets.
First of all, our synthetic datasets demonstrate strong standalone performance ($S_1$ and $S_2$) on most test sets when compared to target domain training data.
We also show cross-domain performances of \doqa{} and \quac{} on each other for comparison, where we observe their training sets to lag well behind our synthetic data, which was also not generated specifically for any of the two domains.

A closer look at the answerable (\textsc{a}) and unanswerable (\textsc{ua}) class performances reveals the same disparity between $S_1$ and $S_2$ as before (\eg{}, in Table~\ref{table:intrinsic-analysis}): $S_1$ performs better on answerable queries and $S_2$ on unanswerable ones.
We therefore also consider a $50$$:$$50$ mixture of the two as a third training set ($10,000$ conversations), termed $S_3$ in the table, to find out how well they complement each other.
The results confirm the strength of their combination, as the mixture outperforms the individual datasets on all test sets.

Our final \sft{} evaluation measures the ability of our synthetic data to augment in-domain gold data, for which we fine-tune the model trained on $S_3$ -- the best synthetic dataset -- further on target domain training examples separately for \doqa{} and \quac{}.
As indicated by the results in Table ~\ref{table:sft-results} (last $2$ rows), this augmentation provides a strong boost to results in both domains over training only on  gold data, with improvements ranging from $10.1\%$ to $13.9\%$.
These remarkable results showcase the out-of-domain utility of our synthetic data derived from only $6$ simple Wikipedia-based demonstrations.
We discuss other methods of augmentation that we experimented with in Appendix~\ref{Appendix: SFT training details}.

\section{Related Work}
\label{section:related-work}

% Scratchpad \cite{nye2021show} showed that LLMs can be trained to imitate a thought process to solve tasks in a step by step manner, greatly benefiting performance. (see also wei2022chain 's review of other older works)

%Chain-of-Thought (CoT) prompting \cite{wei2022chain}, showed that few-shot prompting \cite{brown2020language} can be used to induce a model to produce intermediate thought steps and this increases overall performance.
Chain-of-thought (\chot) prompting \cite{wei2022chain} showed that \llm{}s can be few-shot prompted \cite{brown2020language} to produce intermediate reasoning steps, or ``thoughts'', which can improve their performance. 
%Prompts were case specific and manually created (see \cite[Appendix G]{wei2022chain}. 
Prompts were created manually and were application-specific (see \cite[Appendix G]{wei2022chain}). 
%Chain of Verification (CoVe) \cite{dhuliawala2023chain} introduces instead a \textit{structured} CoT prompting approach to reduce hallucinations and shows the benefit of separating the though space into stages with different prompts.
Chain-of-verification (\textsc{c}o\textsc{v}e) \cite{dhuliawala2023chain} adopted a structured approach to \chot{} for answer verification and error correction, splitting the thought process into verification planning and execution steps that utilize their own prompts, finally producing a rewritten response with reduced hallucination.
%Also related to this, the recitation-augmented CoT approach \cite{sun2022recitation}, recite related knowledge stored in the model parameters before answering the question. 
Also related is the recitation-augmented \chot{} approach of \citet{sun2022recitation}, which recites related knowledge stored in the \llm{}'s parameters before answering a question. 
%Other structured CoT prompting approaches are ReAct \cite{yao2023react}, ReWoo \cite{xu2023rewoo}, which apply it to LLM-powered agents, and interleave though steps with use of external tools e.g. a search engine.
Other structured \chot{} prompting work include \textsc{r}e\textsc{a}ct \cite{yao2023react} and \textsc{r}e\textsc{w}oo \cite{xu2023rewoo}, which leverage \llm{}-powered agents to solve complex tasks through interleaved steps, with use of external tools such as search engines. 

%The proposed approach is a type of structured CoT and is closest to CoVe and recitation CoT. Compared to these approaches we focus in multi-turn grounded generation, utilizing thought steps specific for this scenario i.e. evidence finding. Most importantly, we prove the usefulness of this technique for synthetic data generation instead of direct CoT use at test time.
The proposed approach draws inspiration from the above studies, but aims to solve the complex task of multi-turn content-grounded conversation generation, focusing on the key associated requirements of determining question answerability and knowledge selection.
More importantly, we produce synthetic data that can train agents for direct inference, removing the need for computationally expensive \chot{} prompting during inference.

%Regarding imitation and generation of chain-of-thought, the CoT precursor Scratchpad \cite{nye2021show}, showed that LLMs can be trained to imitate a thought process to solve tasks in a step by step manner and the benefit of generating synthetic CoT data for training. It did not however use prompting and was limited to code-like or abstract tasks. 

%When considering prior work, is is also important to consider reproducibility. CoT, recitation, ReAct, ReWoo and Scratchpad experiments use closed source models (GPT-3.5\footnotemark\footnotetext{Which is being deprecated and is no longer acessible \url{https://platform.openai.com/docs/deprecations/instructgpt-models}}, PaLM \cite{chowdhery2023palm}, Codex\cite{chen2021evaluating}, Lambda). This makes reproducibility difficult and obscures the factors of success, since some of the models lack detailed descriptions. Here we use open source models for which the weights are available (Falcon\cite{penedo2023refinedweb}, FLAN \cite{tay2022ul2}, LLaMA-2 \cite{touvron2023llama}).

The \chot{} precursor Scratchpad \cite{nye2021show} showed that models can also be fine-tuned to execute thought processes in order to solve a task in a step-by-step manner, and the benefits of generating synthetic \chot{} data for training. It did not, however, use prompting and was limited to relatively close-ended problems. Constitutional AI \cite{bai2022constitutional} used a two-step \chot{} process (critique and revise) to generate corrected responses. Like this work, and unlike Scratchpad, it removed the \chot{} and kept only the final response to generate the synthetic data. Unlike our work, Constitutional AI focuses on reducing harmfulness and uses this step as part of a bigger process that includes reinforcement learning.

Finally, it is worth noting that many prior works relied on closed-source models such as \textsc{gpt}-3.5\footnotemark\footnotetext{Which is being deprecated and is no longer accessible \url{https://platform.openai.com/docs/deprecations/instructgpt-models}}, \textsc{p}a\textsc{lm} \cite{chowdhery2023palm}, Codex \cite{chen2021evaluating} and \textsc{l}a\textsc{mda} \cite{thoppilan2022lamda}, and are thus not reproducible.
Here we only use open-source models whose weights are publicly available, both for generation: \textsc{Falcon} \cite{penedo2023refinedweb} and \textsc{Flan} \cite{tay2022ul2}, and for evaluation: \textsc{Mixtral} \cite{jiang2024mixtral}, \textsc{Llama-}$2$ \cite{touvron2023llama} and \textsc{Falcon}.

\section{Conclusion}
\label{section:conclusion}
We introduce a structured chain-of-thought (\schot{}) prompting approach to generating multi-turn content-grounded conversations and empirically show that by relying on minimal human effort to create only six seed conversations, we can produce high-quality synthetic data.
Designated states for hallucination mitigation and the use of supporting tools enable our methods to generate agent utterances that are highly faithful to grounding documents.
Used as training data, our generated conversations train high-performance models as evaluated on out-of-domain test sets, successfully augmenting target-domain human-labeled data.
Future work will explore more complex conversational settings, \eg{}, multi-document grounding and response generation for harder, more ambiguous questions.

\clearpage

\section*{Limitations}
\label{section:limitations}
The goal of this work is to reduce hallucination in pre-trained \llm{}s through structured \chot{} prompting and \lm{} augmentation, improving overall generation quality.
Even though we provide ample empirical evidence of successful hallucination mitigation through both intrinsic and extrinsic evaluation, given the high cost of manual labor involved in evaluating long multi-turn conversations generated by a large number of algorithms, we only perform automatic evaluation, relying on cutting-edge \llm{}s and factual consistency checking models.
While the use of only $6$ seed conversations was useful in demonstrating the strength of our approach, from the perspective of supervised fine-tuning, our generated data could be more diverse and train even better models if more demonstrations were utilized.

\begin{comment}
\section*{Acknowledgements}

This document has been adapted
by Steven Bethard, Ryan Cotterell and Rui Yan
from the instructions for earlier ACL and NAACL proceedings, including those for 
ACL 2019 by Douwe Kiela and Ivan Vuli\'{c},
NAACL 2019 by Stephanie Lukin and Alla Roskovskaya, 
ACL 2018 by Shay Cohen, Kevin Gimpel, and Wei Lu, 
NAACL 2018 by Margaret Mitchell and Stephanie Lukin,
Bib\TeX{} suggestions for (NA)ACL 2017/2018 from Jason Eisner,
ACL 2017 by Dan Gildea and Min-Yen Kan, 
NAACL 2017 by Margaret Mitchell, 
ACL 2012 by Maggie Li and Michael White, 
ACL 2010 by Jing-Shin Chang and Philipp Koehn, 
ACL 2008 by Johanna D. Moore, Simone Teufel, James Allan, and Sadaoki Furui, 
ACL 2005 by Hwee Tou Ng and Kemal Oflazer, 
ACL 2002 by Eugene Charniak and Dekang Lin, 
and earlier ACL and EACL formats written by several people, including
John Chen, Henry S. Thompson and Donald Walker.
Additional elements were taken from the formatting instructions of the \emph{International Joint Conference on Artificial Intelligence} and the \emph{Conference on Computer Vision and Pattern Recognition}.
\end{comment}

\bibliography{custom}

\appendix

\section{Prompts}
\label{appendix-section:prompts}

\subsection{Prompts for the Remaining States}
\label{appendix-subsection: prompts-for-states}
\begin{figure*}[ht]
    \centering        \includegraphics[width=.8\linewidth]{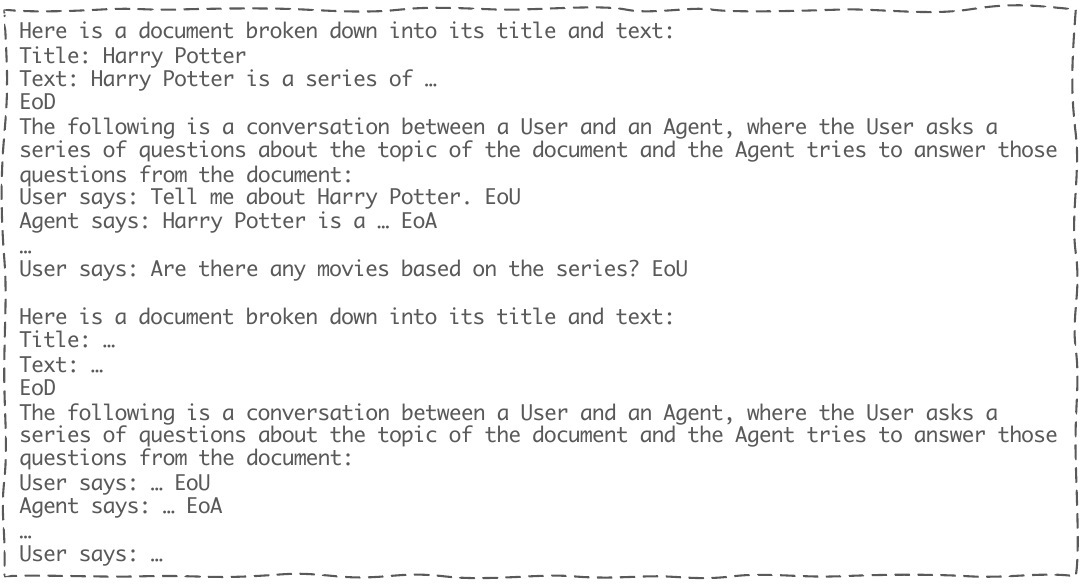}
        \caption{Prompt for a pre-trained \llm{} in state \bolduustate{}: user utterance generation (\S{\ref{section:methodology}}).}
        \label{figure:uu-demo}
\end{figure*}

\begin{figure*}[ht]
    \centering
        \includegraphics[width=.8\linewidth]{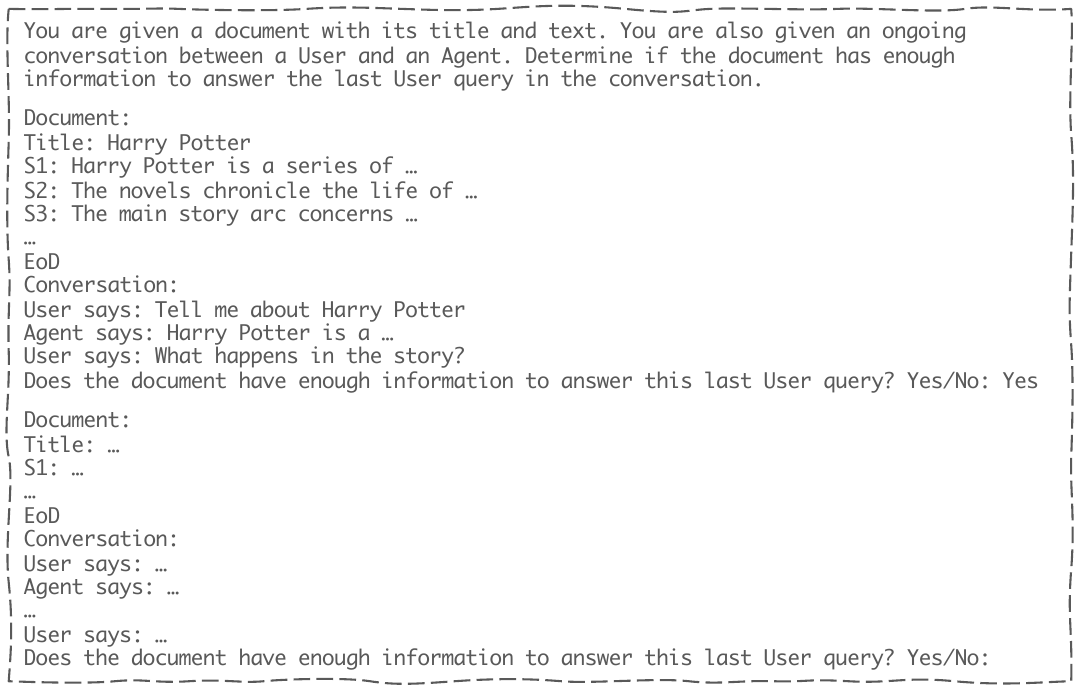}
        \caption{Prompt for an instruction-following \llm{} assistant in state \boldacstate{}: question answerability classification (\S{\ref{section:methodology}}).}
        \label{figure:ac-demo}
\end{figure*}

\begin{figure*}[ht]
    \centering
        \includegraphics[width=.6\linewidth]{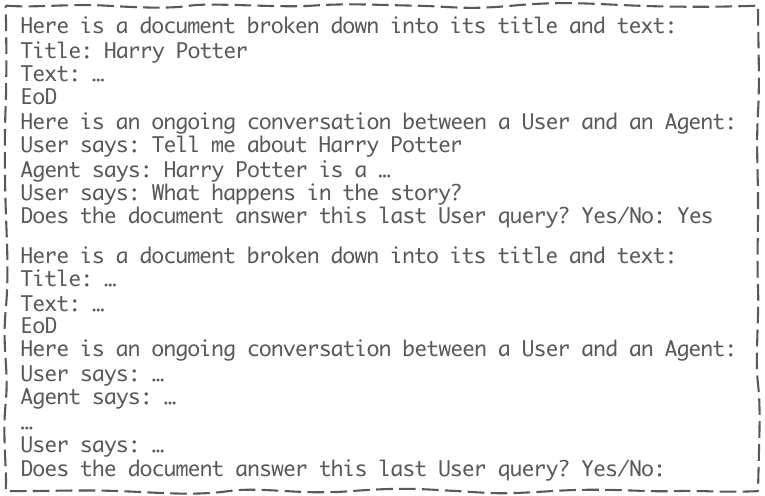}
        \caption{Prompt for a pre-trained \llm{} (\ie{}, no assistant) in state \boldacstate{}: question answerability classification (\S{\ref{section:methodology}}).}
        \label{figure:ac-plm-demo}
\end{figure*}

\begin{figure*}[ht]
    \centering
        \includegraphics[width=.6\linewidth]{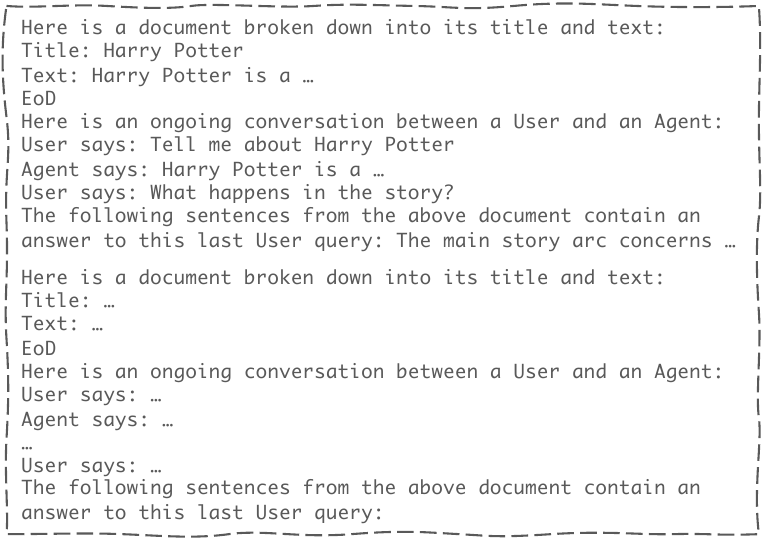}
        \caption{Prompt for a pre-trained \llm{} (\ie{}, no assistant) in state \boldssstate{} (\S{\ref{section:methodology}}).}
        \label{figure:ss-plm-demo}
\end{figure*}

In Figure~\ref{figure:au-ss-demos}, we showed the prompts used in agent utterance generation and answer sentence selection.
Figures~\ref{figure:uu-demo}, \ref{figure:ac-demo}, \ref{figure:ac-plm-demo} and \ref{figure:ss-plm-demo} illustrate our remaining prompts used at various stages of the different algorithms.

\subsection{Details of \icl{} Demonstrations}
\label{appendix-subsection: prompting-details}
In states \bolduustate{} and \boldaustate{}, we use two full document-grounded conversations as demonstrations.
In question answerability classification (\boldacstate{}), we use $3$ positive and $3$ negative exemplars for the \flanultwo{} assistant, and $2$ positive and $2$ negative exemplars for \falcons{}; for the latter, a larger number of demonstrations often exceeds the maximum input length limit.
This is an example of an advantage that comes with \llm{} augmentation (\S{\ref{section:introducion}}), where a different tool can execute actions and/or leverage resources that the primary \llm{} cannot.
In sentence search (\boldssstate{}) we use $6$ exemplars for the \flanultwo{} assistant; for \falcons{}, we are only able to use $3$.

\begin{comment}
\begin{figure*}[ht]
    \centering
        \includegraphics[width=1\linewidth]{latex/images/parts-1-2.png}
        \caption{Left: Agent utterance generation (\boldaustate{}) with a vanilla \llm{} such as \falcons{}.\\
        Right: Knowledge sentence selection (\boldssstate{}) with an instruction-following model such as \flanultwo{}.}
        \label{figure:au-and-ss-demos}
\end{figure*}

\begin{figure}[ht]
    \centering
        \includegraphics[width=1\linewidth]{latex/images/part-1.png}
        \caption{Left: Agent utterance generation (\boldaustate{}) with a vanilla \llm{} such as \falcons{}.}
        \label{figure:au-demo}
\end{figure}

\begin{figure}[ht]
    \centering
        \includegraphics[width=1\linewidth]{latex/images/part-2.png}
        \caption{Knowledge sentence selection (\boldssstate{}) with an instruction-following model such as \flanultwo{}.}
        \label{figure:ss-demo}
\end{figure}    
\end{comment}

% \clearpage

\begin{minipage}{0.9\linewidth}
\begin{lstlisting}[caption={Instruction and Input format},captionpos=b,label={lst:instruction}]
INSTRUCTION = "Given the document and the current conversation between a user and an agent, your task is to generate the next response from the agent.

While generating the agent response you should: Determine if agent response needs information from document.
(a) If yes, generate agent response using only precise information present in the document.
(b) If not, generate CANNOTANSWER. "

input = INSTRUCTION + " \n\n "
input += "Text: " + DOC_TITLE + " : " + DOC_TEXT + "\n\n"
input += f"Input: "

for utt, speaker in zip(dialog["utterances"],dialog["speakers"]):
    input += f"User: " if speaker == "user" else "Agent: "
    input += f"{utt.strip()} "

input = input.rstrip()
input += f"\n\nOutput:"
\end{lstlisting}
\end{minipage}

\section{\sft{} Details}
\label{Appendix: SFT training details}
In this section, we provide details on the instruction and input format used for our SFT experiments. We also present results from all steps from our workflow described in Section \ref{subsubsection:extrinsic-eval-sft-results}.

\subsection{\sft{} Instruction Prompt}
\label{Appendix: SFT Instruction prompt}
We use the instruction and input format described in Listing \ref{lst:instruction} when fine-tuning the \falcont{} model (with \qlora{}).

We use $128$ as the maximum number of tokens for generation during our SFT experiments. We filter out few instances from the \doqa{} and \quac{} datasets based on length to ensure that each input instance is less than $1920$ tokens ($2048$ model max input length - $128$ max output tokens for generation). We create input instances as mentioned above and use the following check:

\begin{lstlisting}
tokenizer = AutoTokenizer.from_pretrained("tiiuae/falcon-40b")
if len(tokenizer(input)['input_ids']) > (2048 - 128):
    return False
else:
    return True
\end{lstlisting}

This filtering reduces size of the \quac{} data as follows:
For the train set, the number of instances is reduced from 76129 to 75194; for the dev set: from 7439 to 7332; and for the test set: from 7354 to 7069 instances.

We fine-tune the \falcont{} model\footnote{tiiuae/falcon-7b} (with \qlora{}) across all training data with 2 Tesla A100 GPUs. We use the following settings across all training runs:
\begin{itemize}
    \item quantization method is set to `fp4'
    \item LoRA rank, alpha, dropout and target modules are set to 8, 32, 0.1 and ["query\_key\_value"]
    \item Batch size and learning rate are set to 1 and $4.0e-4$
    \item DeepSpeedFusedAdam is used as the optimizer with weight decay = 0.1, betas = [0.9, 0.95] and eps = $1e-10$
    \item warmup steps are set to 1000 steps with a linear learning rate schedule.
\end{itemize}

We save checkpoints at periodic intervals during training and evaluate the models on the dev dataset to select the checkpoint that achieves the best validation performance for the final evaluation on the test set. For synthetic data, the models are trained for 15 epochs. For \textsc{GOLD} data, the models are trained for 10 epochs and the augmented models '\textsc{D} augmented w/ \textsc{S3}' and '\textsc{Q} augmented w/ \textsc{S3}' are fine-tuned for 5 epochs.

\subsection{Identifying Best Synthetic Data}
\label{Appendix: Identifying best synthetic data}
As described in Section \ref{subsubsection:extrinsic-eval-sft-results}, we adopt a step-by-step workflow to identify the best synthetic data for the best downstream task performance. In this section, we present the details for each step and share our results from these steps on our two downstream tasks: \doqa{} and \quac{}.

\subsubsection{Step 1: Identifying the best algorithms}
\label{Appendix: Step 1: Identifying best State Transition}

\begin{table*}[h]
    \small
    \centering
    \begin{tabular}{l|ccc|}
    \textbf{Training Set} & \textsc{a} & \textsc{ua} & \textsc{hm} \\
         \hline
         \multicolumn{4}{c}{\doqa{}}\\
         \hline
         $D$: \doqa{} \textsc{Gold} & $24.5$ & $65.5$ & $35.7$\\
         \hline
         \boldtwostatetrans{} & $22.4$ & $55.4$ & $\mathbf{31.9}$ \\
         \boldfourstatetrans{} with a \flanultwo{} Assistant & $12.8$ & $95.6$ & $22.5$ \\
         \boldfourstatetrans{} with no Assistant & $18.9$ & $53.0$ & $27.9$\\
         \boldthreestatetranswithac{} with a \flanultwo{} Assistant
         & $11.7$ & $96.0$ & $20.8$\\
         \boldthreestatetranswithac{} with no Assistant
          & $17.6$ & $73.5$ & $28.4$\\
          \hline
         \multicolumn{4}{c}{\quac{}}\\
         \hline
         $Q$: \quac{} \textsc{Gold} & $25.2$ & $83.0$ & $38.7$ \\
         \hline
         \boldtwostatetrans{} & $25.8$  & $43.0$ & $32.2$ \\
         \boldfourstatetrans{} with a \flanultwo{} Assistant   & $21.7$  & $71.5$ & $\mathbf{33.3}$ \\
         \boldthreestatetranswithac{} with no Assistant  & $26.1$  & $41.0$ & $31.9$ \\
         \boldthreestatetranswithac{} with a \flanultwo{} Assistant   & $15.4$  & $88.5$ &  $26.2$ \\
         \boldfourstatetrans{} with no Assistant   & $16.6$  & $54.0$ & $25.4$ \\
    \end{tabular}
    \caption{Performance of \falcont{} models fine-tuned (with \qlora{}) on 1,000 synthetic conversations on \doqa{} Cooking dev set and \quac{} dev set.}
    \label{table:sft-results-appendix-1000}
\end{table*}
% The above is from Tables 10 & 11 in Jatin's Overleaf

In Step 1, we use the synthetic conversations generated on 1,000 Wikipedia passages from Section \ref{table:intrinsic-analysis} using various state transitions and fine-tune \falcont{} models (with \qlora{}) on them. In Table \ref{table:sft-results-appendix-1000}, we present results of our SFT evaluation using models fine-tuned only on 1000 synthetic conversations. For evaluation, we use the official \doqa{} Cooking dev set and \quac{} dev set.

We observe that among the synthetic datasets, "\boldtwostatetrans{}" model performs best on the \textsc{A} class.
For the \textsc{UA} class, two approaches: "\boldfourstatetrans{} with a \flanultwo{} Assistant" and "\boldthreestatetranswithac{} with a \flanultwo{} Assistant" perform quite well, with the former performing better on the \textsc{A} class. Hence, for our next stage of SFT experiments with larger synthetic data, we choose the "\boldtwostatetrans{}" and the "\boldfourstatetrans{} with a \flanultwo{} Assistant" settings as our preferred synthetic data generation methods for the \doqa{} and \quac{} downstream tasks.

\subsubsection{Step 2: Identifying the best synthetic data mixture}
\label{Appendix: Step 2: Identifying best synthetic data mixture}

In Step 2, we fine-tune models on $10,000$ synthetic conversations ($10x$ from Step 1). In Table \ref{table:sft-results-appendix-10000-50-50-mix}, we present results of our SFT evaluation for these models. We use the \doqa{} and \quac{} dev set as before for evaluation. We observe that with $10,000$ conversations, "$S1$: \boldtwostatetrans{}" model performs similar to $1,000$ conversations in Table \ref{table:sft-results-appendix-1000} for both \doqa{} and \quac{}. For \doqa{}, we notice a significant improvement in the performance of "$S2$: \boldfourstatetrans{} with a \flanultwo{} Assistant" model ($7.5\%$ absolute gain on \textsc{HM} metric).  

In addition to improvements from larger synthetic data size, the most significant finding is with the $S1 \cup S2$ approach where we fine-tune \falcont{} model with a $50-50$ ratio of $S1$ and $S2$. We observe that models trained with $S1 \cup S2$ approach outperform models trained with $S1$ or $S2$ individually. This showcases that the synthetic data generated using our different approaches complements each other and improves the performance on the downstream task.

\begin{table*}[h]
    \small
    \centering
    \begin{tabular}{l|ccc|}
    \textbf{Training Set} & \textsc{a} & \textsc{ua} & \textsc{hm} \\
         \hline
         \multicolumn{4}{c}{\doqa{}}\\
         \hline
         $D$: \doqa{} \textsc{Gold} & $24.5$ & $65.5$ & $35.7$\\
         \hline
         $S1$: \boldtwostatetrans{} & $22.4$ & $59.4$ & $32.5$ \\
         $S2$: \boldfourstatetrans{} with a \flanultwo{} Assistant & $18.7$ & $90.0$ & $31.0$ \\
         $S1 \cup S2$ & $22.1$ & $75.1$ & $34.2$ \\
         \hline
         \multicolumn{4}{c}{\quac{}}\\
         \hline
         $Q$: \quac{} \textsc{Gold} & $25.2$ & $83.0$ & $38.7$ \\
         \hline
         $S1$: \boldtwostatetrans{} &  $24.6$ & $55.4$ & $34.1$ \\
         $S2$: \boldfourstatetrans{} with a \flanultwo{} Assistant & $19.3$ & $84.4$ & $31.4$ \\
         $S1 \cup S2$ & 
         $26.2$ & $58.8$ & $36.2$ \\
    \end{tabular}
    \caption{Performance of \falcont{} models fine-tuned (with \qlora{}) on 10,000 synthetic conversations on \doqa{} Cooking dev and \quac{} dev sets.}
    \label{table:sft-results-appendix-10000-50-50-mix}
\end{table*}
% The above is from Tables 5 & 7 in Jatin's Overleaf

\subsubsection{Step 3: Further improvement using data augmentation}
\label{Appendix: Step 3: Further improvement using data augmentation}

We explore several other approaches for utilizing the synthetic data, such as:
\begin{itemize}
    \item augmenting the synthetic data with \textsc{GOLD} data and fine-tuning from scratch
    \item further fine-tuning the model with \textsc{GOLD} data already trained on synthetic data, and
    \item further fine-tuning the model with synthetic data already trained on \textsc{GOLD} data
\end{itemize}

We use \doqa{} dataset for running these additional experiments and report our results in Table \ref{table:sft-results-appendix-10000-augmentation-doqa}. We observe that the data augmentation improves performance, even when using only $20\%$ of the synthetic data mixture $S3$. Using more synthetic data for augmentation does not yield additional gain in performance. However, this requires retraining the models from scratch, which may be expensive based on model and dataset size. The $S3$ on $D$ model, achieved by further fine-tuning the model (already trained on \doqa{} \textsc{GOLD} data) with $S3$ data does not improve performance. This is expected since our synthetic data is not related to the \textsc{GOLD} data; hence the drop in performance.

The $D$ on $S3$ model, \textit{aka} Alignment model, is achieved by using the model fine-tuned on our synthetic data mixture $S3$ and further fine-tuning it with the \doqa{} \textsc{GOLD} data outperforms the models trained from scratch with data augmentation. This shows that our synthetic data is of good quality and can be used to train a good initial model, which can be used for further domain alignment with \textsc{GOLD} data available for the domain.

\begin{table}[h]
    \small
    \centering
    \begin{tabular}{l|ccc|}
    \textbf{Training Set} & \textsc{a} & \textsc{ua} & \textsc{hm} \\
         \hline
         $D$: \doqa{} \textsc{Gold} & $24.5$ & $65.5$ & $35.7$\\
         \hline
         $S1$: \boldtwostatetrans{} & $22.4$ & $59.4$ & $32.5$ \\
         \shortstack[l]{$S2$: \boldfourstatetrans{}\\\quad w/ \flanultwo{} Asst.} & $18.7$ & $90.0$ & $31.0$ \\
         \hline
         $S3$: $S1 \cup S2$ & $22.1$ & $75.1$ & $34.2$ \\
         \hline
         $D \cup S3 (20\%)$  & 
         $33.0$ & $75.9$ & $46.0$ \\
         $D \cup S3 (40\%)$  & 
         $35.6$ & $62.3$ & $45.3$ \\
         $D \cup S3(60\%)$  & 
         $36.2$ & $63.5$ & $46.1$ \\
         $D \cup S3(80\%)$  & 
         $34.8$ & $68.7$ & $46.2$ \\
         $D \cup S3(100\%)$  & 
         $34.9$ & $67.9$ & $46.1$ \\
         \hline
         $S3$ on $D$ model & 
         $16.2$ & $89.6$ & $27.5$ \\
         $D$ on $S3$ model & 
         $37.7$ & $66.3$ & \textbf{$48.1$} \\
         \hline
    \end{tabular}
    %\caption{Performance of \falcont{} models fine-tuned (with \qlora{}) on 10,000 synthetic conversations on \doqa{} Cooking dev set. $S1$ refers to \boldtwostatetrans{} and $S2$ refers to \boldfourstatetrans{} with a \flanultwo{} Assistant as covered in Table \ref{table:sft-results-appendix-10000-50-50-mix}.}
    \caption{Performance of \falcont{} models fine-tuned (with \qlora{}) on 10,000 synthetic conversations on \doqa{} Cooking dev set.}
    \label{table:sft-results-appendix-10000-augmentation-doqa}
\end{table}

\begin{comment}
    
\begin{table}[h]
    \centering
    \begin{tabular}{l|ccc|}
    \textbf{Training Set} & \textsc{a} & \textsc{ua} & \textsc{hm} \\
         \hline
         $Q$: \quac{} \textsc{Gold} & $25.2$ & $82.9$ & $38.7$ \\
         \hline
         $S3$: $S1 \cup S2$ & 
         $26.2$ & $58.8$ & $36.2$ \\
         \hline
         $S3$ on $Q$ model & 
         $24.7$ & $65.9$ & $35.9$ \\
         $Q$ on $S3$ model & 
         $35.0$ & $72.7$ & $47.3$ \\
          % & 
         % $$ & $$ & $$ \\
    \end{tabular}
    \caption{Performance of \falcont{} models fine-tuned (with \qlora{}) on 10,000 synthetic conversations on \quac{} dev set. $S1$ refers to \boldtwostatetrans{} and $S2$ refers to \boldfourstatetrans{} with a \flanultwo{} Assistant as covered in Table \ref{table:sft-results-appendix-10000-50-50-mix}.}
    \label{table:sft-results-appendix-10000-augmentation-quac}
\end{table}
\end{comment}

\end{document}